\documentclass[12pt]{article}
\usepackage{amsmath}
\usepackage{amssymb}
\usepackage{graphicx,psfrag,epsf}
\usepackage{enumerate}
\usepackage{comment}
\usepackage[table]{xcolor}
\usepackage[ruled,vlined]{algorithm2e}
\DeclareMathOperator{\di}{d\!}
\DontPrintSemicolon
\SetKwFor{For}{for (}{) $\lbrace$}{$\rbrace$}
\usepackage{natbib}
\setcitestyle{authoryear,open={(},close={)}}
\usepackage{hyperref}
\hypersetup{colorlinks,citecolor=red, filecolor=black, linkcolor=blue, urlcolor=black} 
\usepackage{amsthm}
\newtheorem{theorem}{Theorem}

\newcommand{\blind}{0}
\usepackage{caption}
\theoremstyle{plain}
\newtheorem*{theorem*}{Theorem}

\begin{document}

\def\spacingset#1{\renewcommand{\baselinestretch}%
{#1}\small\normalsize} \spacingset{1}

\def\spacingset#1{\renewcommand{\baselinestretch}%
{#1}\small\normalsize} \spacingset{1}


\if0\blind
{
  \title{\bf Deep Learning with Functional Inputs}
  \author{Barinder Thind\hspace{.2cm}\\
    Department of Statistics \& Actuarial Science\\ Simon Fraser University\\
    \\
    Kevin Multani \\
    Department of Physics\\ Stanford University\\
    \\
    Jiguo Cao \\
    Department of Statistics \& Actuarial Science\\ Simon Fraser University\\}
    \date{}
  \maketitle
} \fi

\if1\blind
{
  \bigskip
  \bigskip
  \bigskip
  \begin{center}
    {\LARGE\bf Deep Learning with Functional Inputs}
\end{center}
  \medskip
} \fi
\vspace{-2em}
\bigskip
\begin{abstract}
\spacingset{1.5}
We present a methodology for integrating functional data into deep densely connected feed-forward neural networks. The model is defined for scalar responses with multiple functional and scalar covariates. A by-product of the method is a set of dynamic functional weights that can be visualized during the optimization process. This visualization leads to greater interpretability of the relationship between the covariates and the response relative to conventional neural networks. The model is shown to perform well in a number of contexts including prediction of new data and recovery of the true underlying functional weights; these results were confirmed through real applications and simulation studies. A forthcoming R package is developed on top of a popular deep learning library (\texttt{Keras}) allowing for general use of the approach.

\end{abstract}

\noindent%

\newpage

\spacingset{1.5}
\section{Introduction}
\label{sec:intro}
Functional data analysis (FDA) is a growing statistical field for analyzing curves, surfaces, or any multidimensional functions, in which each random function is treated as a sample element \citep{Ramsay05,ferraty2006nonparametric}. Functional
data is found commonly in many applications such as time-course gene expressions and brain scan images. The ever-expanding umbrella that encompasses deep-learning methodologies has thus far largely excluded the usage of functional covariates. With the advent and rise of functional data analysis, it is natural to extend neural networks and all of their recent advances to the functional space. The main goal of this article is to provide a new means of modelling functional data for scalar response prediction in the form of neural networks. 

Let $Y$ be a scalar response variable, and $X(t)$, $t \in \mathcal{T}$, be the functional covariate. Several models have been proposed to predict the scalar response $Y$ with the functional covariate $X(t)$. For instance, when the scalar response $Y$ follows a normal distribution, the conventional functional linear model is defined as:
\begin{align*}
    E(Y|X) = \alpha + \int_{\mathcal{T}}\beta(t)X(t)\di t,
\end{align*}
where $\alpha$ is the intercept term, and $\beta(t)$ is the functional coefficient which represents the cumulative linear effect of $X(t)$ on $Y$ \citep{cardot1999functional}. This model was extended to the general functional linear model:
\begin{align*}
    E(Y|X) = g\left(\alpha + \int_{\mathcal T}\beta(t)X(t)\di t\right)
\end{align*}
when the scalar response $Y$ follows a general distribution in an exponential family, where $g(\cdot)$ is referred to as the link function and has a specific parametric form for the corresponding distribution of $Y$ \citep{muller2005generalized}. When this link function $g(\cdot)$ has no parametric form, the model is called the functional single index model \citep{jiang2011functional}. Other predictive methods in the realm of FDA are related to various estimation methods of $\beta(t)$. For example, a partial least squares approach was proposed by \cite{preda2007pls} where an attempt was made to estimate $\beta(t)$ such that it maximized the covariance between the response, $Y$ and the functional covariate, $X(t)$. \cite{ferraty2006nonparametric} proposed a non-parametric model:
\begin{align*}
    E(Y|X) = r\left(X(t)\right),
\end{align*}
where $r(\cdot)$ is a smooth non-parametric function that is estimated using a kernel approach. Another model, which serves as an extension to the previous, is the semi-functional partial linear model \citep{aneiros2006semi} defined as:
\begin{align*}
    E(Y|X) = r\left(X(t)\right) + \sum_{j=1}^J w_{j}Z_{j},
\end{align*}
where $Z_j, j = 1,\ldots,J,$ is the scalar covariate that is observed in the usual multivariate case, and the function $r(\cdot)$ is also estimated with no parametric form using kernel methods. 

All of the functional models above have been shown to have some level of predictive success. However, we show that the general neural network in this article outperforms these models. We propose a novel methodology for deep network structures in a general regression framework for longitudinal data. The form of a single neuron $v$, for $K$ functional covariates and $J$ scalar covariates in this model is expressed as:
\begin{align}
\label{eq1}
    v = g\left(\sum_{k = 1}^{K}\int_{\mathcal T} \beta_{k}(t)x_{k}(t)\di t + \sum_{j = 1}^{J}w_{j}z_{j} + b\right),
\end{align}
where $g(\cdot)$ is some activation (i.e. non-linear) function, $w_{j}$ is the weight associated with the scalar covariate $z_{j}$,
and $\beta_{k}(t)$ is the functional weight that corresponds to the functional covariate $x_k(t)$. 


With respect to neural networks, we have seen a growing number of approaches, some of which have resulted in previous benchmarks being eclipsed. For example, \cite{krizhevsky2012imagenet} won the ImageNet Large-Scale Visual Recognition Challenge in 2012 improving on the next best approach by an over 10\% increase in accuracy. \cite{rossi2005functional} proposed a neural network in which they used a single functional variable for classification problems. \cite{he2016deep} introduced residual neural networks, which allowed for circumvention of vanishing gradients -- an innovation that carved a path for networks with exponentially more layers thus further improving error rates. These successes however, have come at the cost of interpretability. As the models become more complex, it becomes an increasingly difficult task to make sense of the network parameters. On the other hand, conventional linear regression models have a relatively clear interpretation of the parameters estimates \citep{seber2012linear}. In the functional linear regression case, the coefficient parameters being estimated are functions $\beta_k(t), k=1,\ldots,K$, rather than a set of scalar values. This paper details an approach that makes the functional coefficient traditionally found in the regression model readily available from the neural network process in the form of functional weights; the expectation is that this increases the interpretability of the neural networks while maintaining the superior predictive power.

Our paper has three major contributions. First, we introduce the general framework of functional neural networks (FNNs) with a methodology that allows for deep architectures with multiple functional and scalar covariates, the usage of modern optimization techniques \citep{kingma2014adam, ruder2016overview}, and hyperparameters such as early stopping \citep{yao2007early} and dropout \citep{srivastava2014dropout}, along with some justifications that underpin the approach. Second, we introduce functional weights that are smooth functions of time and can be much easier to interpret than the vectors of parameters estimated in the usual neural network. This is exemplified in our applications and simulations. Finally, branching off this work is a forthcoming R package developed on top of a popular deep learning library (\texttt{Tensorflow/Keras}) that will allow users to apply the proposed method easily on their own data sets.

The rest of our paper is organized as follows. We first introduce the methodology for functional neural networks in \hyperref[sec:meth]{Section 2}. Additionally, commentary is provided on the interpretation potential, weight initialization, and the hyperparameters of these networks. Then, \hyperref[sec:results]{Section 3} provides results from real world examples; this includes prediction comparisons among a number of methods for multiple data sets. In \hyperref[sec:sim]{Section 4}, we use simulation studies for the purpose of recovering the true underlying coefficient function $\beta_k(t)$, and to test the predictive accuracy of multivariate and functional methods in four different contexts. Lastly, \hyperref[sec:conc]{Section 5} contains some closing thoughts and new avenues of research for this kind of network.

\section{Methods}
\label{sec:meth}

\subsection{Functional Neural Networks}

We will begin with a quick introduction of traditional neural networks which are made up of hidden layers each of which contains some number of neurons. Let $n_{u}$ be the number of neurons in the $u$-th hidden layer. Each neuron in each layer is some non-linear transformation of a linear combination of each activation in the previous layer. An activation value is the output from each of these neurons. For example, the first hidden layer $\boldsymbol{v}^{(1)}$ would be defined as $\boldsymbol{v}^{(1)} = g\left(\boldsymbol{W}^{(1)}\boldsymbol{x} + \boldsymbol{b}^{(1)}\right),$ where $\boldsymbol{x}$ is a vector of $J$ covariates, $\boldsymbol{W}^{(1)}$ is an $n_1$ x $J$ weight matrix, $\boldsymbol{b}^{(1)}$ is the intercept (often referred to as the bias in machine learning texts), and $g(\cdot)$ is some activation function that transforms the resulting linear combination \citep{ESL}. The choice of the function $g:\mathbb{R}^{n_1} \rightarrow \mathbb{R}^{n_1} $ is highly context dependent. The rectifier \citep{hahnloser2000digital} and sigmoidal functions \citep{han1995influence} are popular choices for $g$. Note that the vector $\boldsymbol{x}$ corresponds to a single observation of our data set. The resulting vector $\boldsymbol{v}^{(1)}$ is $n_{1}$-dimensional. This vector contains the activation values to be passed on to the next layer. 

Thus far, the assumption has been that $\boldsymbol{x}$ is $J$ -dimensional. However, we wish now to consider the case when our input is infinite dimensional defined over some finite domain $\mathcal{T}$, i.e., we postulate that our input is a functional covariate $x(t): \mathcal{T} \rightarrow \mathbb{R}$, $t\in\mathcal{T}$. By finite domain we mean that $a<t<b$ for $a,b \in \mathbb{R}$. We must weigh this functional covariate at every point along its domain. Therefore, our weight must be infinite dimensional as well. We define this weight as $\beta(t)$. The form of a neuron with a single functional covariate in the first layer then becomes
\begin{align}
\label{eq:2}
    v^{(1)}_{i} = g\left(\int_{\mathcal T}\beta_{i}(t)x(t)\di t + b_{i}^{(1)}\right),
\end{align}
where the subscript $i$ is an index that denotes one of the $n_{1}$ neurons in this first hidden layer, i.e., $i\in \lbrace{1,2,\dots,n_1\rbrace}$. We omit the superscript on the functional weight $\beta(t)$, because this parameter only exists in the first layer of the network. 


 The functional weight $\beta_i(t)$ is expressed as a linear combination of basis functions: $\beta_i(t) = \sum_{m = 1}^{M}c_{im}\phi_{im}(t) = \boldsymbol{c}_i^T \boldsymbol{\phi}_i(t)$, where $\boldsymbol{\phi}_i(t) = (\phi_{i1}(t), \ldots, \phi_{iM}(t))^T$ is a vector of basis functions, and $\boldsymbol{c}_i = (c_{i1}, \ldots, c_{iM})^T$ is the corresponding vector of basis coefficients. The basis coefficients for $\beta_i(t)$ will be initialized by the network; these initializations will then be updated as the network learns. Common choices of basis functions are the B-splines and the Fourier basis functions \citep{Ramsay05}. We also note that the evaluation of the neuron in \hyperref[eq:2]{Equation~(2)} results in some scalar value. This implies that the rest of the $u - 1$ layers of the network can be of any of the usual forms (feed-forward, residual, etc.). Using these basis approximations of $\beta_i(t)$, we can simplify to get that the form of a single neuron is:
\begin{align}
\label{eq:single_neuron}
    v_{i}^{(1)} &= g\left(\int_{\mathcal T} \beta_{i}(t)x(t)\di t + b_{i}^{(1)}\right)\nonumber\\
    &= g\left(\int_{\mathcal T} \sum_{m=1}^{M} c_{im}\phi_{im}(t)x(t)\di t + b_{i}^{(1)}\right)\nonumber\\
    &= g\left(\sum_{m=1}^{M} c_{im}\int_{\mathcal T}\phi_{im}(t)x(t)\di t + b_{i}^{(1)}\right),
 \end{align}
\noindent where the integral in \hyperref[eq:single_neuron]{Equation~(3)} can be approximated with numerical integration methods such as the composite Simpson's rule \citep{Suli03}.

\begin{figure}[htbp]
  \centering
  \includegraphics[scale = 0.54]{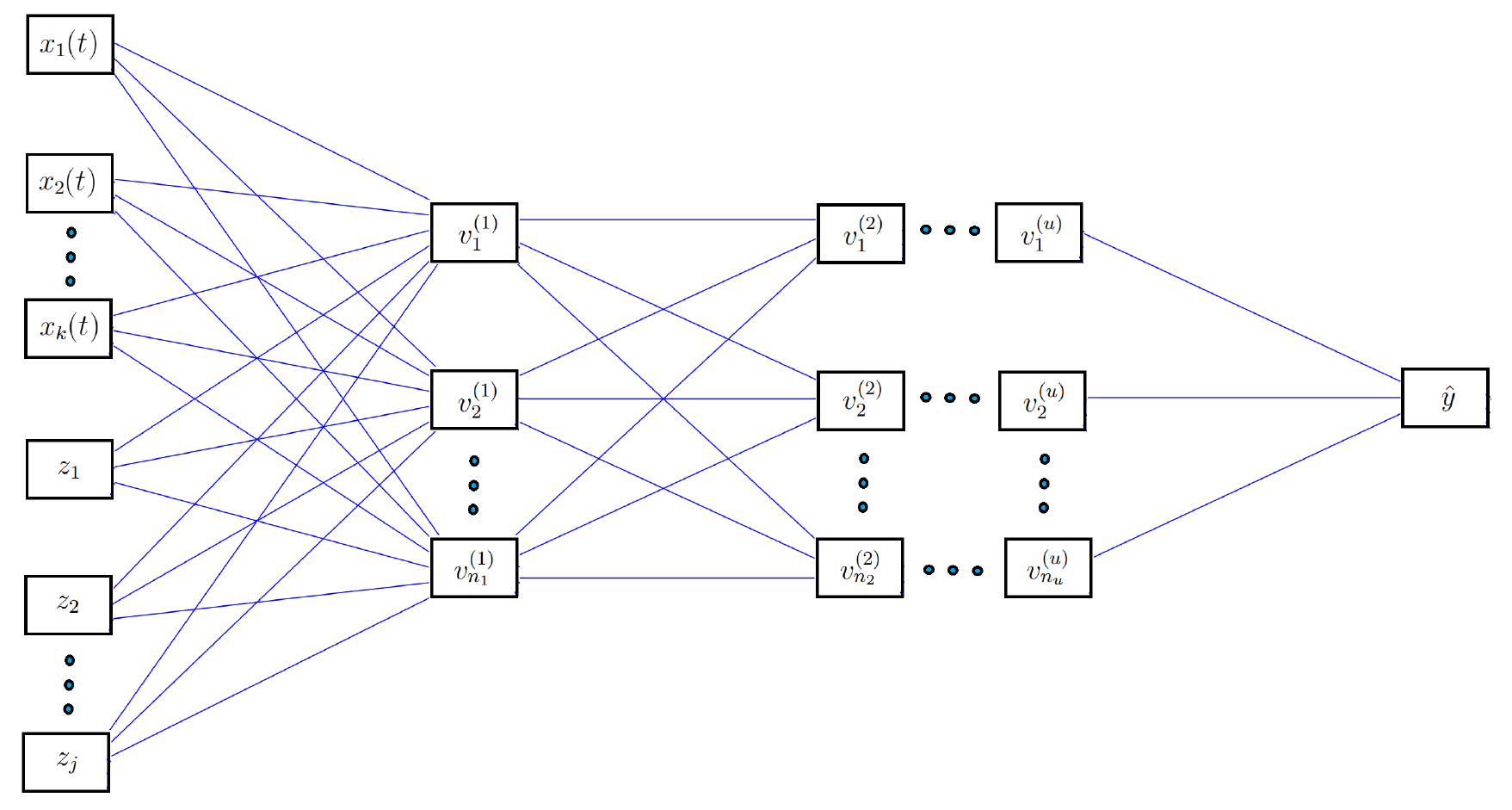}
  \label{fig:schema_fnn}
  \spacingset{1}
  \caption{Schematic of a general functional neural network for when the inputs are functions, $x_k(t)$, and scalar values, $z_j$. The response/output of this network is a scalar value, $\hat{y}$. }
\end{figure}

We can now consider the generalization for $K$ functional covariates and $J$ scalar covariates. Consider the input layer as presented in \hyperref[fig:schema_fnn]{Figure 1}. The covariates correspond to the $\ell$-th observation can be seen as the set:
\begin{align*}
    \text{input}_\ell = \{x_{1}(t), x_{2}(t), ..., x_{K}(t), z_{1}, z_{2}, ..., z_{J}\}.
\end{align*}
Then, the $i$-th neuron of the first hidden layer corresponding to the $\ell$-th observation can be formulated as (we suppress the index, $\ell$ because this expression does not change with the observation number):
\begin{align*}
    v_i^{(1)} = g\left(\sum_{k = 1}^{K}\int_{\mathcal T} \beta_{ik}(t)x_{k}(t)\di t + \sum_{j = 1}^{J}w_{ij}^{(1)}z_{j} + b_i^{(1)}\right),
\end{align*}
where
\begin{align*}
    \beta_{ik}(t) = \sum_{m = 1}^{M}c_{ikm}\phi_{ikm}(t),
\end{align*}
$\phi_{ikm}(t)$ is the basis function and $c_{ikm}$ is the corresponding basis coefficient. This neuron formulation is the core of this methodology as alluded to in \hyperref[eq1]{Equation~(1)}. Note that $c_{ikm}$ here is unique at the initialization for each functional weight, $\beta_{ik}(t)$ -- the choice of these initializations is discussed later in the article. Also, in this formation, we have assumed that $M$ is the same across all $K$ functional weights; it could be the case that the user prefers some functional weight to be defined using a different number of basis functions than $M$ say $M_{k}$, so this is left as a hyperparameter.

Having specified the form, we define the following general formation of the first layer:
\begin{align*}
    v_i^{(1)} &= g\left(\sum_{k = 1}^{K}\int_{\mathcal T} \sum_{m = 1}^{M_k}c_{ikm}\phi_{ikm}(t)x_{k}(t)\di t + \sum_{j = 1}^{J}w^{(1)}_{ij}z_{j} + b^{(1)}_i\right)\\
    &= g\left(\sum_{k = 1}^{K}\sum_{m = 1}^{M_k}c_{ikm}\int_{\mathcal T} \phi_{ikm}(t)x_{k}(t)\di t + \sum_{j = 1}^{J}w^{(1)}_{ij}z_{j} + b^{(1)}_i\right).
\end{align*}
To justify this, we consider the one-layer case and consider Theorem 1 in \cite{cybenko1989approximation}, which states that linear combinations of the form $G(x) = \sum_{n=1}^N \alpha_n \sigma\left(y_n^Tz + \theta_n\right)$ exhibit the quality that, under some conditions, the 1-norm between the function you want to learn $f(x)$, and the function $G(x)$ can be arbitrarily small. Since we are looking at the one-layer case, we have $n_1$ neurons (indexed from $i = 1$ to $n_1$) and we omit the superscript that indexes the layer number, $u=1$. Additionally, we fix the observation number $\ell$ because it does not play a role in the proof (you can apply the same argument to each observation).

\begin{theorem}
    Let $g: \mathbb{R} \to \mathbb{R}$ be any continuous sigmoidal function, $I_{n}$ denote the $n-$ dimensional hypercube $[0, 1]^{n}$ and $\mathcal{C}(I_{n})$ denote the space of continuous functions. Then, the finite sum of the following form, is dense in $\mathcal{C}(I_n)$:
    \begin{align*}
        h(t) = \sum_{i=1}^{n_1} \Psi_{i} g\left(\sum_{k = 1}^{K}\left(\int_{\mathcal{T}} \beta_{ik}(t)x_{k}(t)\di t\right) + \sum_{j = 1}^{J}w_{ij}z_{j} + b_{i}\right),
    \end{align*}
    meaning that for any $f(t) \in \mathcal{C}(I_{n})$ and for $\epsilon > 0$, the function $h(t)$ obeys:
    \begin{align*}
        |h(t) - f(t)| < \epsilon.
    \end{align*}
\end{theorem}

A proof is provided in the supplementary document. After running through this set of initial neurons in the first layer and calculating the activations for the layers following, we can arrive at a final value. The output will be single dimensional. In order to assess performance, we can use some loss function, $R$; for example, the mean squared error
\begin{align*}
     R(\theta) = \sum_{\ell = 1}^{N}\left(y_\ell - \hat{y}_\ell(\theta)\right)^{2},
\end{align*}
where $\theta$ is the set of parameters defining the neural network, $y_{\ell}, \ell=1,\ldots,N,$ is the observed data for the scalar response, and $\hat{y}_\ell$ is the output from the functional neural network. 

\subsection{Functional Neural Network Training}

Having defined the general formation of functional neural networks, we can now turn our attention to the optimization of this kind of network. We will consider the usual backpropogation algorithm \citep{rumelhart1985learning}. While in the implementation, we used an \texttt{adam()} optimizer \citep{kingma2014adam}, we can explain the general process when the optimization scheme uses stochastic gradient descent.

Given our generalization and reworking of the parameters in the network, we can note that the set $\theta^{'}$ making up the gradient associated with the parameters is:
\begin{align*}
    \theta^{'} = \bigg\{\bigcup_{k = 1}^{K} \bigcup_{m = 1}^{M_{k}} \bigcup_{i = 1}^{n_{1}} \dfrac{\partial R}{\partial c_{ikm}}, \bigcup_{u = 1}^{U} \bigcup_{j = 1}^{J_{u}} \bigcup_{i = 1}^{n_{u}} \dfrac{\partial R}{\partial w_{iju}}, \bigcup_{u = 1}^{U} \bigcup_{i = 1}^{n_{u}} \dfrac{\partial R}{\partial b_{iu}}\bigg\}.  
\end{align*}
This set exists for every observation, $\ell$. We are trying to optimize for the entirety of the training set, so we will move slowly in the direction of the gradient. The rate at which we move, which is called the learning rate, will be denoted by $\gamma$. For the sake of efficiency, we will take a subset of the training observations, which is called a mini-batch, for which we calculate $\theta^{'}$. Then, letting $\Bar{a} = \sum_{\ell=1}^{N_{b}} {a_{\ell}^{'}}/{N_{b}}$, where $a^{'}_{\ell} = \partial R/\partial a_{\ell}$ is the derivative of any parameter $a \in \theta$ for the $\ell$-th observation and $N_{b}$ is the size of the mini-batch. The update for $a$ is $ a = a - \gamma \Bar{a}$ \citep{ruder2016overview}. This process is repeated until all partitions (mini-batches) of the data set are completed thus completing one training iteration; the number of training iterations is left as a hyperparameter. We summarize the entire network process in \hyperref[algo:a]{Algorithm 1}.

Lastly, we would like to emphasize that the number of parameters in the functional neural network presented here has decreased significantly under this approach. Consider a longitudinal data set where we have $N$ observations and $P$ scalar repeat measurements of some covariate at different points along its continuum. Passing this information into a network will mean that the number of parameters in the first layer will be $(P + 1)\cdot n_1$. Note that in our network, the number of parameters in the first layer is a function of the number of basis functions we use to define the functional weight. The number of basis functions $M$, we use to define this functional weight will be less than $P$ as there is no need to have a functional weight that interpolates across all our observed points -- we prefer a smooth effect across the continuum to avoid fitting to noise. Therefore, good practice indicates that the number of parameters in the first layer of our network is $(M + 1)\cdot n_1$ where $M < P$.



\begin{algorithm}[h!]
\spacingset{1.2}
    \vspace{1em}
  \KwIn{Functional and Scalar Observations}
  \KwOut{$\theta$}
  -----------------------------------------------------------------------------------------------------------\\
  1. Set Hyperparameters:\\ \hspace{1em}$\gamma$, \# of Basis Functions, Activation Functions, \# of Layers, \# Of Neurons per Layer,\\ \hspace{1em} Training Iterations, Loss Function\\
  2. Initialize weights of network, $\theta_{p}$\\
  3. \textbf{for} $q$ in 1:\textit{Training Iterations}\\
  \hspace{1em} 3i. Forward Pass\\
  \hspace{3em} a. Observed data passed to first hidden layer\\
  \hspace{3em} b. Approximate $\int_{\mathcal{T}} \phi_{km}(t)x(t)\di t$ = $\tilde{\phi}_{km}$ for each basis function, $m$ and for\\\hspace{3em} each functional covariate, $k$\\ 
  \hspace{3em} c. Calculate $g\left(\sum_{k = 1}^{K}\sum_{m = 1}^{M_k}c_{km}\tilde{\phi}_{km} + \sum_{j = 1}^{J}w^{(1)}_{j}z_{j} + b^{(1)}\right)$ for each neuron\\     
  \hspace{3em} d. Pass activations in c. to any other network architecture as per usual\\
  \hspace{3em} e. Calculate loss: $R(\theta)$\\
  \hspace{1em} 3ii. Backward Pass\\
  \hspace{3em} a. Compute $\theta^{'}$\\
  \hspace{3em} b. $\forall$ $a$ $\in$ $\theta_{p}$, update $a$ as: $a = a - \gamma\Bar{a}$\\
  \hspace{1em} 3iii. \textbf{If} $q \leq$ \textit{Training Iterations}\\
  \hspace{3em} a. Go to 3i.\\
  \hspace{3em} b. \textbf{Else:} Go to 4.\\
  4. Return $\theta = \theta_{p}$
  \vspace{1em}

  \caption{Functional Neural Networks}
  \label{algo:a}
\end{algorithm}

\subsection{Functional Weights}

\begin{figure}[h!]
  \centering
  \includegraphics[scale = 0.47]{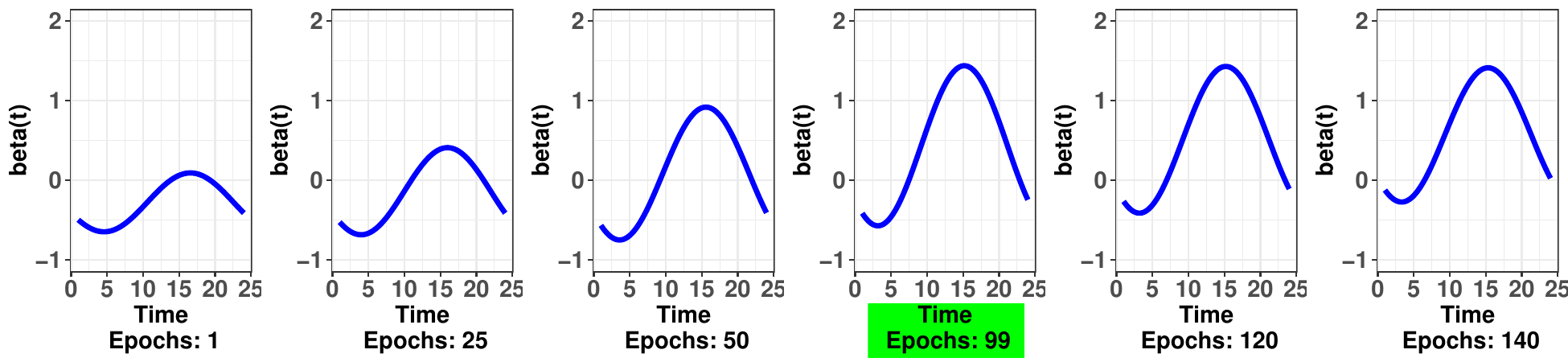}
  \label{fig:change_of_curve}
  \spacingset{1}
  \caption{An example of how the functional weight changes over the number of training iterations or epochs as they are referred to in neural network literature. }
  
\end{figure}

Since a leading contributor to the black-box reputation of neural networks is the inordinate amount of changing weights and intercepts, it would be helpful to consider rather a function defined by these seemingly uninterpretable numbers. In a functional neural network, we are estimating functional weights  $\beta_{ik}(t) = \sum_{m = 1}^{M}c_{ikm}\phi_{ikm}(t)$. These functional weights are akin to the ones predicted in the functional regression model \citep{Ramsay05}; the final set of functional weights $\beta_{ik}(t)$ can be compared with the one estimated from a function linear model. In the case of multiple neurons, we take the average of the estimated functional weight $\hat{\beta}_{k}(t) = \sum_{i=1}^{n_1}\hat{\beta}_{ik}(t)/{n_1}$. Over iterations of the network, as it is trained, we can see movement of the functional weight over its domain. Since these parameters can be visualized, it can be much easier to garner intuition about the relationship between the functional covariates and the response. \hyperref[fig:change_of_curve]{Figure 2} may be illuminating. At the 99th training iteration, the validation error stops decreasing with respect to some threshold. We can see that the difference in the curves is most pronounced in the beginning and is least pronounced after the model finds some local extrema. In this example, the functional weights were initialized from a uniform distribution but a more drastic change in the shape could be seen with a different initialization and a different choice of basis functions for the functional weight.

  

\subsection{Weight Initialization and Parameter Tuning}

For any usual neural network, the weights and intercepts can be initialized in a number of ways. For example, in \cite{kim1991weight} weights are initialized based on a boundary that allowed for quicker convergence. Another approach is to consider a zero-initialization, i.e., letting the initial parameter values be 0. Many of these approaches have also been compared to one another using various guidelines \citep{fernandez2001weight}. In the case of the networks presented here, this is left as a hyperparameter. Since the implementation is built on top of the \texttt{Keras} architecture \citep{chollet2015keras}, the initialization is dependant on the type of connected layers, but generally the \texttt{glorot\_uniform()} initializer is the choice for dense layers.

Due to the sheer number of hyperparameters in the network, a tuning approach can be used to find optimal values in our applications. The tuning method is to take a list of possible values for each parameter and run a cross-validation \citep{ESL} for all combinations. The number of folds to use depends on the size of the problem. The general scheme is that the function creates a grid, and calculates the K-fold cross-validated mean squared prediction error $\text{MSPE} = \sum_{k=1}^K \sum_{l \in S_k}^N\left(\hat{y}_{l}^{(-k)} - y_{l}\right)^2/N$, where $S_k$ is the $k$-th partition of the data set, and $\hat{y}_{l}^{(-k)}, l \in S_k,$ is the predicted value for $y_l$ by training the functional neural network using the rest of the $K-1$ partitions of the data set. The number of data points in $S_k$ depends on the number of folds. The final output of the tuning function is the combination of hyperparameters that have the minimum value of this cross-validated error. A list of hyperparameters is given in Table S1 in the supplementary materials. One important parameter in this particular kind of network is the number of basis functions that govern the functional weights. Tuning this is fairly important as the number of terms significantly impacts the potential for interpretability and restricts us to some particular shape of the curve. In the examples to come, we tune our hyperparameters using the tuning function in our forthcoming package.

\section{Applications}
\label{sec:results}

\subsection{Bike Rental Data}

An important problem in rental businesses is the amount of supply to keep on-site. If the company cannot meet demands, they are not being as profitable as they can be. If they exceed the required supply, they have made investments that are not yielding an acceptable return. Using the bike rental data set \citep{fanaee2014event}, we look to model the relationship between the total number of daily rentals (a scalar value) and the hourly temperature throughout the day (functional observation). It makes intuitive sense for temperature to be related to the number of bike rentals: on average, if it's cold, less people are likely to rent than if it were warm. We also expect there to be a temporal effect of temperature -- if we have the same temperature at 1pm and 9pm, we would expect more rentals at 1pm under the assumption that less people are deciding to bike later at night. In total, we have data for 102 Saturdays from which we did our analysis -- we chose the same day of the week to eliminate the day-to-day variation. Our functional observations of temperature (to be passed into the network) are formed using 31 Fourier basis functions.

We are first concerned with the accuracy of our predictions. Using $R^{2} = 1 - \sum_{l=1}^N(y_l - \hat{y_{l}})^{2}/\sum_{l=1}^N(y_{l} - \Bar{y})^{2}$ and a 10-fold cross-validated mean squared prediction error, we can compare results for a number of models. Here, we compare with the usual functional linear model, an FPCA approach \citep{cardot1999functional}, a non-parametric functional linear model \citep{ferraty2006nonparametric}, and a functional partial least squares model \citep{preda2007pls}. The results are summarized in \hyperref[table:one]{Table 1}. For the final model, we had a four-layer network with exact hyperparameter configurations being found in Table S2 in the supplementary materials. We observe that FNNs outperform all the other models using both criteria but note that the penalized partial least squares approach and the principal component ridge regression performed comparably.

\begin{center}
\setlength{\tabcolsep}{40pt}
\setlength{\columnseprule}{0.3pt}
\scalebox{0.58}{
\label{table:one}
 \begin{tabular}{|| c | c | c ||} 
 \hline
 \texttt{Model} & \texttt{MSPE}$_{CV}$ & \texttt{$R^{2}$} \\ [1ex] 
 \hline\hline
 Functional Linear Model (Basis) & 0.0723 & 0.515 \\ [1ex]
  \hline\hline
  Functional Non-Parametric Regression & 0.143 & 0.154 \\ [1ex] 
 \hline\hline
 Functional PC Regression & 0.0773 & 0.503 \\ [1ex] 
 \hline\hline
 Functional PC Regression (2nd Deriv Penalization) & 0.128 & 0.0481 \\ [1ex] 
 \hline\hline
 Functional PC Regression (Ridge Regression)  & 0.0823 & 0.464 \\ [1ex] 
 \hline\hline
 Functional Partial Least Squares & 0.0755 & 0.458 \\ [1ex] 
 \hline\hline
 Functional Partial Least Squares (2nd Deriv Penalization) & 0.0701 & 0.545 \\ [1ex] 
  \hline\hline
 Functional Neural Networks & 0.0669 & 0.582 \\ [1ex]
 \hline
\end{tabular}}\\
\spacingset{1}    
\captionof{table}{The $10$-fold cross-validated mean-squared predication error (\texttt{MSPE}$_{CV}$) and $R^2$ of eight models, including the functional neural network (FNN). The FNN performs the best with respect to both measures.}
\end{center}


\begin{figure}[htbp]    
  \centering
  \includegraphics[scale = 0.5]{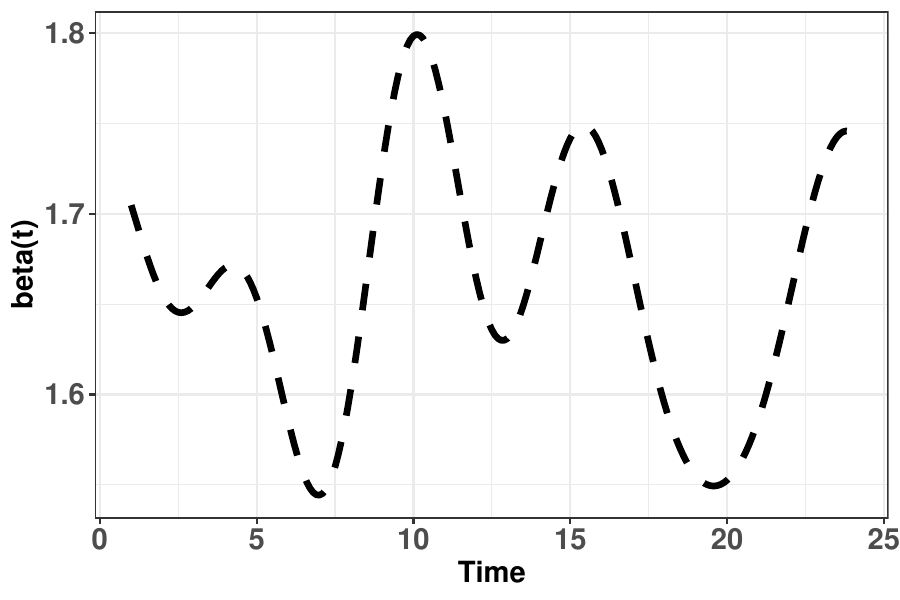}
  \includegraphics[scale = 0.5]{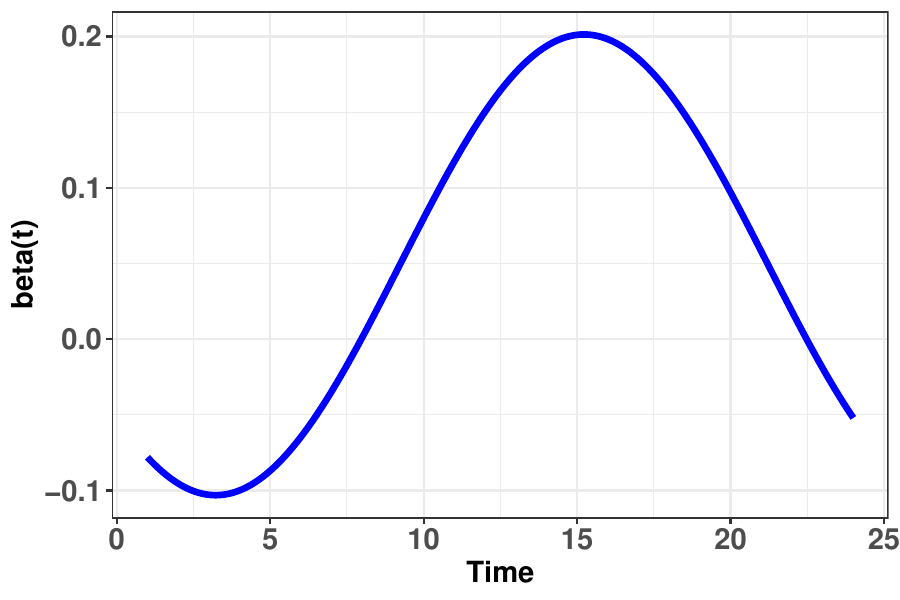}
  \label{fig:fnc_lm_com_bike}
  \spacingset{1}
  \caption{The estimated functional weight $\hat{\beta}(t)$ in the conventional functional linear model (dashed black curve) and the functional neural network (solid blue curve) for the bike rental data set.}
\end{figure}

We can also look to see what the determined relationship is according to the functional linear model and the functional neural network between hourly temperature and daily rentals as indicated by $\beta(t)$.  \hyperref[fig:fnc_lm_com_bike]{Figure 3} shows the estimated weight function $\hat{\beta}(t)$. The optimal number of basis functions was eleven for the functional linear model and three for the functional neural network. For the functional linear model, we note that there seems to be no obvious discernable relationship between hourly temperature and bike rentals. In the case of the functional neural network, we see that there seems to be a positive relationship as we move into the afternoon and that this relationship tapers off as the day ends. We would also expect there to be no effect for when bike rental retailers would be closed, and this is much better reflected in the functional weight from the neural network than the functional coefficient in the functional linear model. We observe different scales for the two and we posit that this difference can be explained by the fact that the functional neural network has a large number of additional parameters that may be explaining some of the variation in the response. Moreover, even though the scale is different, the range of the scale is the same so the relative effect across the continuum is similar. 

\subsection{Tecator Data}

We consider the classic Tecator data set \citep{thodberg2015tecator}. The data are recorded on a Tecator Infratec Food and Feed Analyzer using  near-infrared light (wavelength is $850 \text{ nm - }1050$ $\text{ nm}$) to analyze the samples. Each sample contains meat with different moisture, fat, and protein contents. The goal is to predict the scalar value of fat contents of a given meat sample using the functional covariate of the near infrared absorbance spectrum and the scalar covariate associated with the water contents. Absorbance spectroscopy measures the fraction of incident radiation absorbed by the sample. Samples with higher water composition may exhibit different spectral features (absorbance bands) than samples with higher protein content. Since we are working with functional covariates, we also have access to their derivatives; because this is fundamentally a problem in physics, the derivative information can serve as an important predictor and is used as such. 
 
In total, there are 215 absorbance curves. We used 29 Fourier basis functions to estimate the functional observations. The first 165 absorbance curves are used as the training set, and the predictions are made on the remaining -- this test/train paradigm comes from \cite{febrero2012statistical}; they fitted several models using this paradigm. We present their results along with results from the functional neural network, in \hyperref[table:two]{Table 2}. In the original paper, the authors use the metric $\text{MEP} = \text{ MSPE}/{\text{Var}(y)}$, where MSPE is the average squared errors of the test set and $\text{Var}(y)$ is the variance of the observed response (we can think of $\text{MEP}$ as a rescaling of the $\text{MSPE})$ to assess the models. They also used $R^2$, which we tabulate in \hyperref[table:two]{Table 2}. In the functional neural network, we tuned to find that a six-layer network was optimal with a total of 4029 parameters. Our model has the lowest $\text{MEP}$, but is about $3\%$ lower than the best $R^2$. Most other models perform worse with the Semi-Functional Partial Linear Model \citep{aneiros2006semi} being the most comparable. \smallskip

\begin{center}
\setlength{\tabcolsep}{45pt}
\setlength{\columnseprule}{0.4pt}
\scalebox{0.55}{
\label{table:two}
 \begin{tabular}{||c | c | c ||} 
 \hline
 \texttt{Model} & \texttt{MEP} & \texttt{$R^{2}$} \\ [1ex] 
 \hline\hline
 fregre.basis(X.d1, Fat) & 0.0626 & 0.928 \\ [1ex] 
  \hline\hline
 fregre.basis.cv(X.d2, Fat)  & 0.0566 & 0.965 \\ [1ex] 
 \hline\hline
 fregre.pc(X.d1, Fat) & 0.0580 & 0.950 \\ [1ex] 
 \hline\hline
 fregre.pc(X.d2, Fat) & 0.0556 & 0.954 \\ [1ex] 
 \hline\hline
 fregre.pls(X.d1, Fat)  & 0.0567 &  0.951 \\ [1ex] 
 \hline\hline
 fregre.pls(X.d2, Fat) & 0.0487 & 0.962 \\ [1ex] 
 \hline\hline
 fregre.lm(Fat, X.d1 + Water) & 0.0097 & 0.987 \\ [1ex] 
  \hline\hline
 fregre.lm(Fat, X.d2 + Water) & 0.0119 & 0.986 \\ [1ex] 
  \hline\hline
 fregre.np(X.d1, Fat) & 0.0220 & 0.987 \\ [1ex] 
  \hline\hline
 fregre.np(X.d2, Fat) & 0.0144 & 0.996 \\ [1ex] 
  \hline\hline
 fregre.plm(Fat, X.d1 + Water) & 0.0090 & 0.996 \\ [1ex] 
  \hline\hline
 fregre.plm(Fat, X.d2 + Water) & 0.0115 & 0.997 \\ [1ex] 
  \hline\hline
 FNN(Fat, X.d2 + Water) & 0.00883 & 0.965 \\ [1ex]
 \hline
\end{tabular}}\\
\spacingset{1}    
\captionof{table}{The mean square error of prediction (MEP) and $R^{2}$ of our method and those used in \cite{febrero2012statistical} for analyzing the Tecator data set. The objects X.d1 and X.d2 refer to the first and second derivatives of the near infrared absorbance spectrum curve, respectively.}
\end{center}

Note that we only presented the results using the second derivative of the spectrum curves as the functional covariate because it was the better performer when compared to the network made using the first derivative or the raw functional observations themselves. We did not use multiple functional covariates here because all the other models only used one functional covariate; however, we did use water as a scalar covariate.

\subsection{Canadian Weather Data}

The data set used here has information regarding the total amount of precipitation in a year and the daily temperature for 35 Canadian cities. We are interested in modelling the relationship between annual precipitation and daily temperature. Generally, you would expect that lower temperatures would indicate higher precipitation rates. However, this is not always the case. In some regions, the temperature might be very low, but the inverse relationship with rain/snow does not hold. Our goal is to see whether we can successfully model these anomalies relative to other methods.

The functional observations are defined for the temperature of the cities for which there are 365 (daily) time points, $t$. In total, there are 35 functional observations and the scalar response is the average precipitation across the year. A Fourier basis expansion was used with 65 basis functions defining each of the 35 cities \citep{fda_matlab}. The details of the network hyperparameters is in Table S2 in the supplementary materials. The results from two criteria ($R^{2}$ and a leave-one-out-cross-validated MSPE) are measured for a number of models. We see that the FNN model outperforms all other approaches including the usual neural networks. \hyperref[table:three]{Table 3} summarizes the predictive results. 

\begin{figure}[h!]
  \centering
  \includegraphics[scale = 0.65]{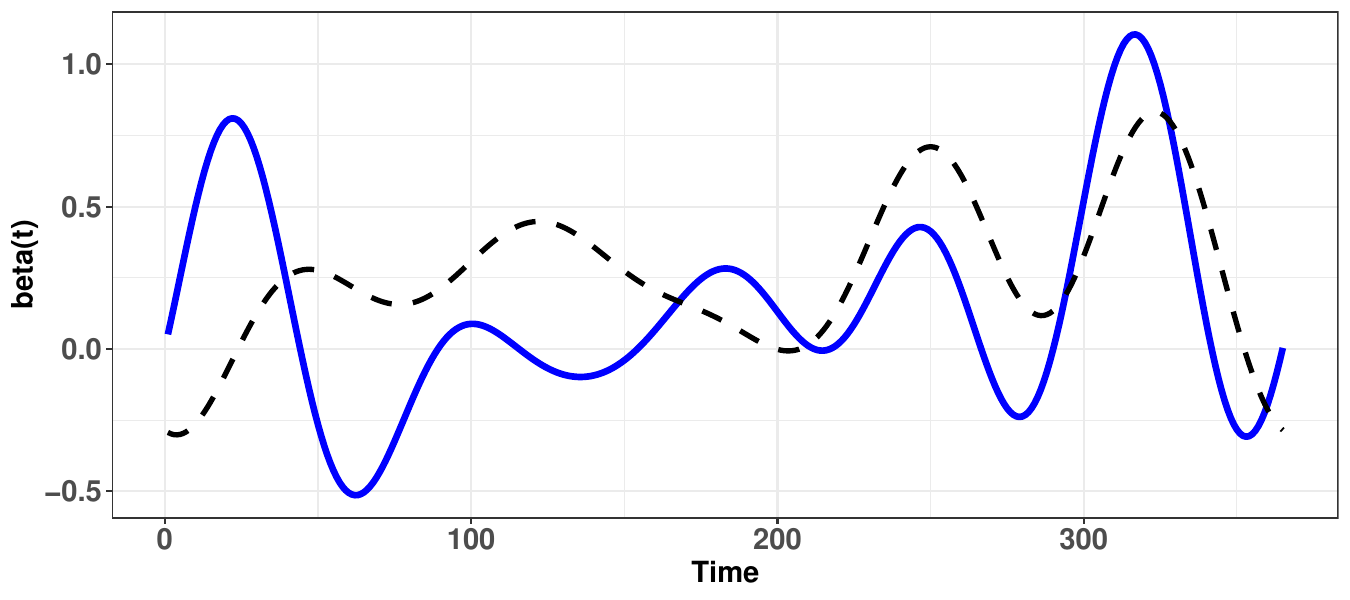}
  \label{fig:fnc_lm_comp}
  \spacingset{1}    
  \caption{The estimated functional weight for the conventional functional linear model (dashed black curve) and the functional neural network (solid blue curve) for the weather data set. }
\end{figure}

We can compare the estimated functional weight from the functional linear model with the functional neural network in \hyperref[fig:fnc_lm_comp]{Figure 4}. For this data set, we decided to keep the number of basis functions the same across both models -- the choice for this was eleven and comes from \cite{fda_matlab}. This was to measure how similar the functional coefficients would be under the same conditions. We observe similar patterns between the two models especially over the second half of the domain. Note that the difference between the two recovered functional weights only accounts for some of the difference in $R^{2}$. The functional neural network has many more parameters allowing for more flexibility in the modelling process and thus the great increase in accuracy. 


\begin{center}
\setlength{\tabcolsep}{45pt}
\setlength{\columnseprule}{0.4pt}
\scalebox{0.64}{
\label{table:three}
 \begin{tabular}{||c | c  | c ||} 
 \hline
 \texttt{Model} & \texttt{MSPE$_{CV}$} & \texttt{$R^{2}$} \\ [1ex] 
 \hline\hline
 Functional Linear Model (Basis) & 0.123 & 0.00312 \\ [1ex] 
  \hline\hline
  Functional Non-Parametric Regression & 0.0647 & 0.0506 \\ [1ex] 
 \hline\hline
 Functional PC Regression & 0.0272 & 0.352 \\ [1ex] 
 \hline\hline
 Functional PC Regression (2nd Deriv Penalization) & 0.0930 & 0.00298 \\ [1ex] 
 \hline\hline
Functional PC Regression (Ridge Regression) & 0.0259 & 0.382 \\ [1ex] 
 \hline\hline
 Functional Partial Least Squares & 0.0449 & 0.177 \\ [1ex] 
 \hline\hline
 Functional Partial Least Squares (2nd Deriv Penalization) & 0.0483 & 0.155 \\ [1ex] 
  \hline\hline
 Neural Networks & 0.126 & 0.0453 \\ [1ex] 
  \hline\hline
Functional Neural Networks & 0.0194 & 0.541 \\ [1ex] 
 \hline
\end{tabular}}\\
\spacingset{1}    
\captionof{table}{The leave-one-out cross-validated mean-squared predication error (MSPE$_{CV}$) and $R^2$ of nine models for the weather data set. }
\end{center}

\section{Simulation Studies}
\label{sec:sim}
\subsection{Recovery of Functional Weight}
\label{sec:4.1}
In this section, we present results from when we know the true underlying functional weight. The goal is to compare the functional weight in the functional neural network to the conventional functional linear model. Useful results here would go a long way in showing that the functional neural network is not only useful for prediction, but can be a valiant tool when the goal is to approximate relationships via parameter estimation. In order to measure this, the integrated mean square error (IMSE) is used, which is defined as
\begin{align*}
    \text{IMSE} = \frac{1}{|\mathcal{T}|}\int_{\mathcal T}(\beta(t) - \hat{\beta}(t))^2 \di t,
\end{align*}
where $\hat{\beta}(t)$ is the estimated functional weight either from the FNN or from the functional regression. We use the following to generate our response:
\begin{align}
\label{eq:sim}
    y_l^{*} = g\left(\alpha + \int_{\mathcal T}\beta(t)x_l(t)\di t\right) + \epsilon_l^{*},
\end{align}
where our choice for $\beta(t)$ is
\begin{align*}
    \beta(t) = m_{1} + m_{2}\sin(t\pi) + m_{3}\cos(t\pi) + m_{4}\sin(2t\pi) +  m_{5}\cos(2t\pi),
\end{align*}
and $\epsilon^\ast$ is sampled from the Gaussian distribution, $\mathcal{N}(0, 1)$. The true functional covariate $x_l(t)$ is generated, depending on the simulation scenario, either from $a\cdot sin(a) + b$ or $c\cdot\exp(a) + sin(a) + b$, where $a$ $\sim \mathcal{N}(0, 1)$, $b$ $\sim \mathcal{N}(0, \frac{l}{100})$, and $c$ $\sim \mathcal{N}(0, 1)$ are parameters that govern the difference between the functional observations.

This generative procedure will be used for four different simulations. In all four, we generate 300 observations randomly using \hyperref[eq:sim]{Equation~(4)} by varying $a$, $b$ and $c$. The coefficients for $\beta(t)$, $m_{1}, ..., m_{5}$, are set beforehand. We fit the functional linear model and the functional neural network for each simulation data set. We cross-validate over a grid for $\lambda$ in order to find a smooth estimate of $\beta(t)$ from the functional linear model. The difference is measured using IMSE. The simulation is replicated 250 times. 

The first simulation is for when the link function $g(\cdot)$ is the identity function. Here, we would expect the functional linear model to perform comparably than the functional neural network due to its deterministic nature and the linear relationship. In the second simulation, we look to see if our method can recover $\beta(t)$ for when the link function is exponential. The third simulation explores this behaviour for a sigmoidal relationship. And lastly, we simulate a logarithmic relationship between the response and the functional covariates. All scenarios except simulation 2 use $c\cdot\exp(a)$ +  $sin(a)$ + $b$ for the data generation. These simulations are summarized as follows:
\begin{align*}
    \label{eq:lol}
    \text{Simulation} \hspace{0.3em} 1: y^{*} &= \alpha + \int_{\mathcal T}\beta(t)x(t)\di t + \epsilon^{*} \\ 
    \text{Simulation} \hspace{0.3em} 2: y^{*} &= \exp\left(\alpha + \int_{\mathcal T}\beta(t)x(t)\di t\right) + \epsilon^{*}\\
        \text{Simulation} \hspace{0.3em} 3: y^{*} &= \frac{1}{1 + \exp{\left(\alpha + \int_{\mathcal T}\beta(t)x(t)\di t\right)}} + \epsilon^{*}\\
    \text{Simulation} \hspace{0.3em} 4: y^{*} &= \log\left(\left|\alpha + \int_{\mathcal T}\beta(t)x(t)\di t\right|\right) + \epsilon^{*}.
\end{align*}

\begin{figure}[htb!]
  \centering
  \includegraphics[scale = 0.38]{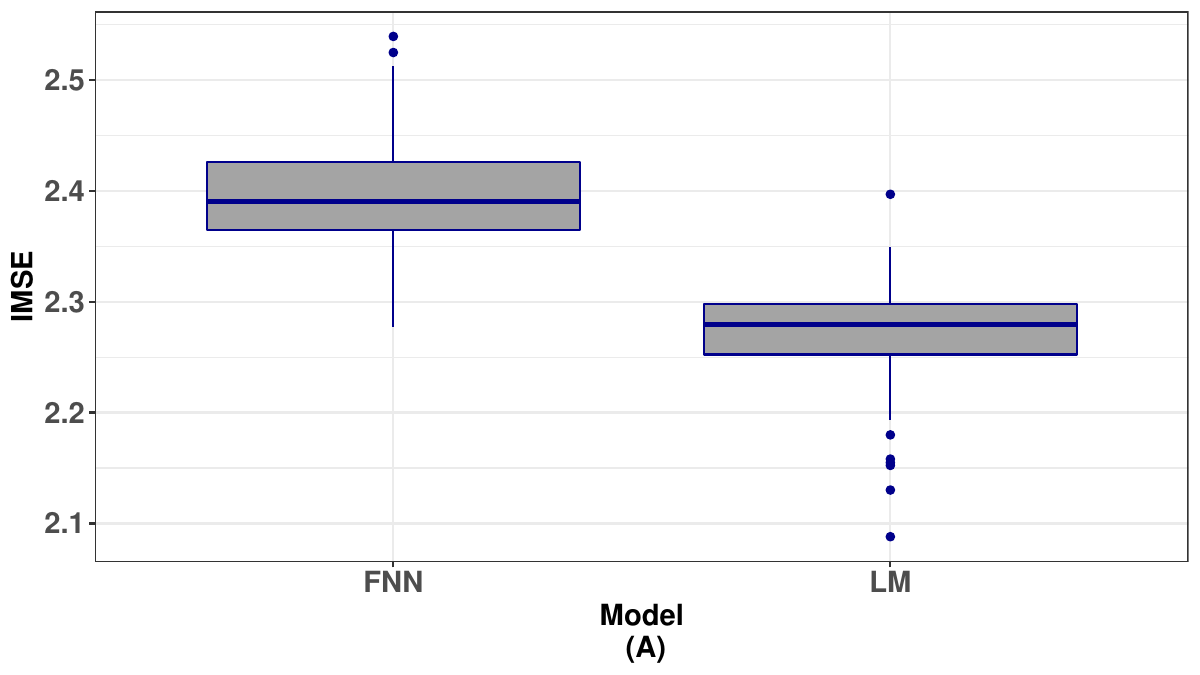}
  \includegraphics[scale = 0.38]{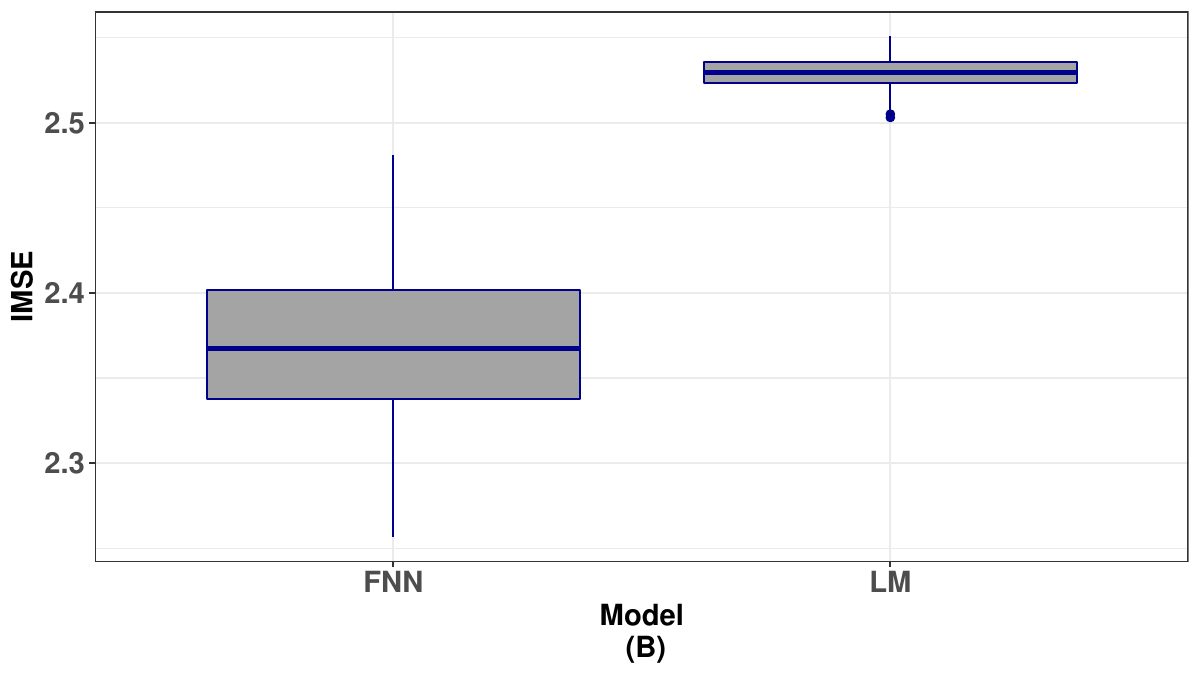}
  \includegraphics[scale = 0.38]{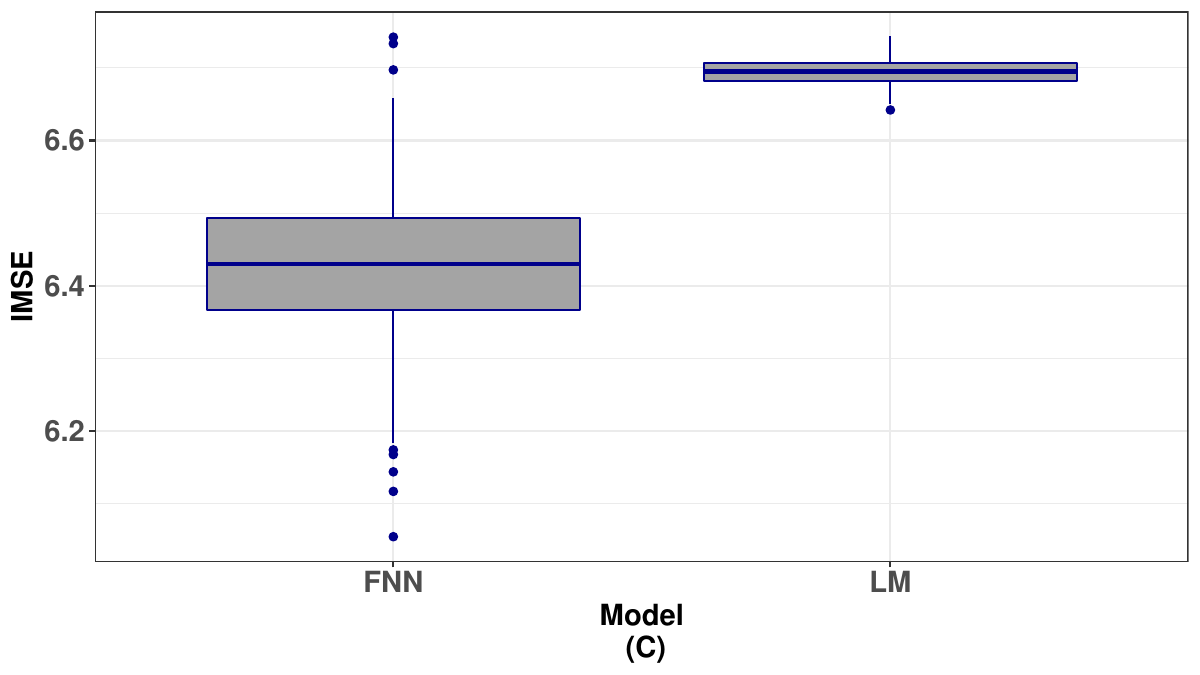}
  \includegraphics[scale = 0.38]{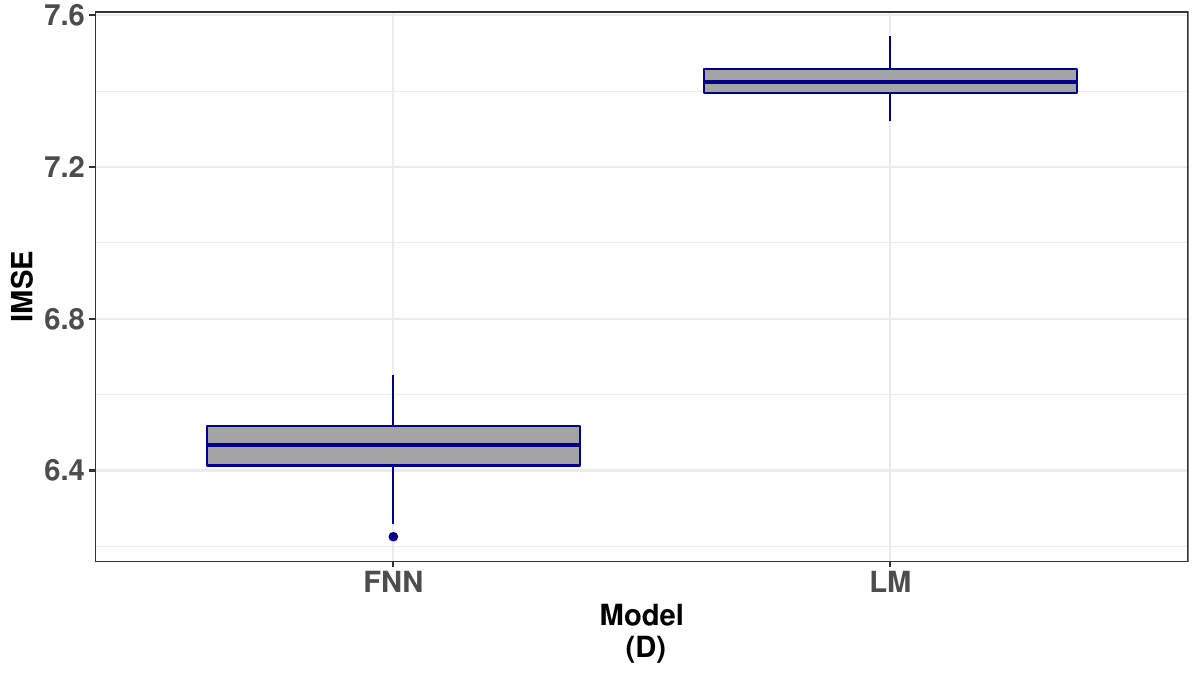}
  \label{fig:sim}
  \spacingset{1}
  \caption{Boxplots of square root of the integrated mean square errors (IMSEs) over 250 simulation replicates for four scenarios. Plot (A) is for when we use the identity link function. Plot (B) are results from the exponential link. The bottom plots, (C) and (D) are the results from simulation 3 and 4, respectively.}
\end{figure}

In all but the fourth scenario, we use a three-layer network with rectifier \citep{hahnloser2000digital} and linear activation functions. In the final scenario, we use a one-layer network with a sigmoidal activation function. With respect to the linear model, we cross-validate over a grid to find the optimal $\lambda$ parameter to smooth the resulting functional weight. 

In \hyperref[fig:sim]{Figure 5}, we present the results for these four simulations. We observe that the usual linear model seems to perform better when the relationship is linear. There are far more parameters in the FNN that are contributing to the prediction of $y$. When the relationship is non-linear, the functional linear model struggles where relatively, the FNN does a much better job in recovering $\beta(t)$.

The averages of these results along with computation times are provided in \hyperref[table:four]{Table 4}. We observe that the functional neural network, as expected, takes a longer time to run across all simulation scenarios. However, we also observe, as the box plots indicate, that for non-linear simulation scenarios, the functional neural network outperforms the functional linear model. This difference seems to be the most pronounced in simulation scenario 4. We also note that because of the stochastic nature i.e., random weight initialization of functional neural networks, there is a higher variance in our estimates when compared with the deterministic functional linear model. We note that with a more rigorous tuning of the functional neural networks, we could further improve these results.

\begin{center}
\setlength{\tabcolsep}{25pt}
\setlength{\columnseprule}{0.4pt}
\scalebox{0.6}{
\label{table:four}
\begin{tabular}{|| c | c | c | c | c | c | c ||}

\hline
              & \multicolumn{3}{c|}{\texttt{Functional  Linear Model}}                                                  & \multicolumn{3}{c|}{\texttt{Functional Neural Networks}}                                                \\ \hline\hline
              & \multicolumn{1}{c|}{Mean}                            & \multicolumn{1}{c|}{SD}     & Avg. Comp. Time & Mean                                                 & SD     & \multicolumn{1}{l|}{Avg. Comp. Time} \\ \hline\hline
Simulation: 1 &  2.27 & \multicolumn{1}{c|}{.0370} & 0.232s   & 2.39                                                & .0476 & 4.68s                         \\ \hline\hline
Simulation: 2 & 2.53                                                & .00901                     & 0.232s   &  2.36 & .0453 & 4.67s                         \\ \hline\hline
Simulation: 3 & 6.70                                                & .0182                      & 0.247s   &  6.43 & .108  & 6.00s                         \\ \hline\hline
Simulation: 4 & 7.43                                                & .0464                      & 0.258s   &  6.46 & .0752 & 6.30s                         \\ \hline
\end{tabular}}\\
\spacingset{1}
\captionof{table}{The square root of the mean along with the associated standard deviation (SD) of the integrated mean square errors (IMSEs) over 250 simulation replicates are provided for both the functional neural networks and the functional linear model. The computation times given are the average per simulation replication.}
\end{center}

\subsection{Prediction}

In this section, we look to see how relatively well our method does under the four different simulation scenarios detailed previously when the task is prediction. That is, we are interested in seeing how FNNs perform versus functional and multivariate approaches. The multivariate methods to be compared include: least squares regression (MLR), LASSO \citep{tibshirani1996regression}, random forests (RF) \citep{breiman2001random}, gradient boosting approaches (GBM, XGB) \citep{friedman2001greedy} \citep{chen2015xgboost}, and projection pursuit regression (PPR) \citep{friedman1981projection}. 

\begin{figure}[!htbp]

  \includegraphics[scale = 0.5]{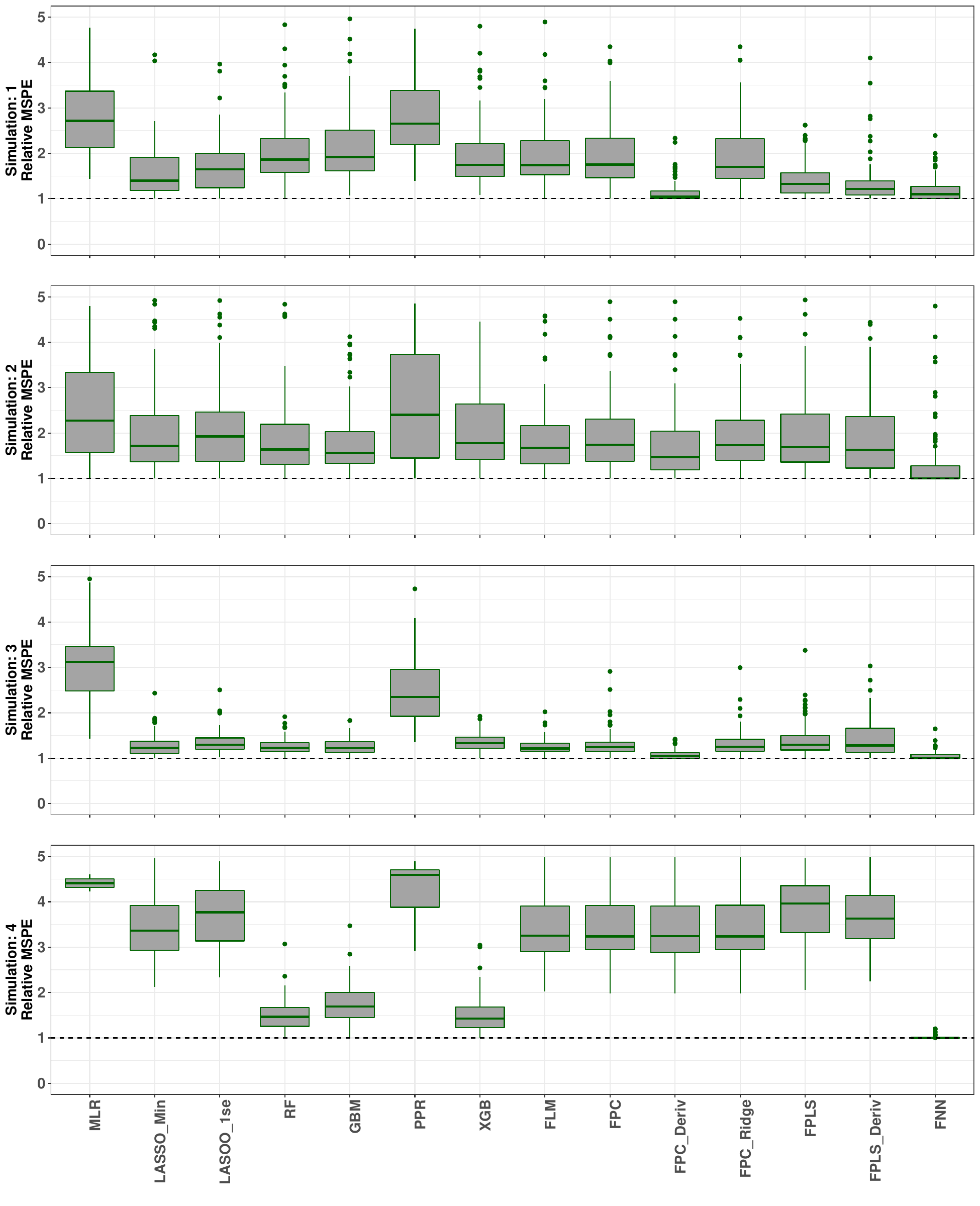}
  \label{fig:sim2}
  \spacingset{1}
  \vspace{-1.5em}
  \caption{Boxplot of the relative mean squared predicted errors (rMSPEs) for fourteen methods in four simulation scenarios.}
\end{figure}

We did not tune our functional neural networks because we found that they performed well in our initial tests irregardless of the tuning. We did make an effort to tune all the other models. For example, the choice of $\lambda$ for the LASSO was made using cross-validation. The tree methods including RF, GBM, and XGB were tuned across a number of their hyperparameters such as the node size, and for PPR, we built models with different numbers of terms and picked the model with the lowest MSPE. In these simulations, we did 100 replicates. For each simulation replicate, we generated 300 functional observations in accordance to \hyperref[eq:sim]{Equation~(4)}. After the realization, we split the data randomly, built a model on the training set, and predicted on the test set. This process is repeated for the same four simulation scenarios as given in  \hyperref[sec:4.1]{Section 4.1}. 

The box plots in \hyperref[fig:sim2]{Figure 6} measure the relative error in each simulation replicate. We call this the relative MSPE defined as: 
\begin{align*}
 \text{rMSPE} = \frac{\text{MSPE}}{\underset{\text{all models}}{\min}\text{MSPE}}.
\end{align*}
For example, on any given simulation replicate, we calculate the MSPE values for each model, and then divide each of them by the minimum in that run. The best model according to this measure will have a value of 1. Error values greater than 1 implies a worse performance. Table S3 in the supplementary materials contains the absolute MSPE values.  


 The relative measure we use makes it easy to compare each model within a simulation, and across the four simulation scenarios. Notably, we see in \hyperref[fig:sim2]{Figure 6} that the functional neural network performs well. This can be attributed to the addition of the functional information passed into the network; the functional neural network can take into account the temporal trend of the data to learn more about the underlying relationship in each training iteration. Therefore, we gain a model that better estimates the underlying true relationship between the covariates and the response in comparison with multivariate approaches.

 In the curve building process, we assume that there is noise associated with the observed discrete values -- by reverse engineering into an approximation of the curve via basis functions, we effectively reduce that noise and then later, when we build the model, we avoid some of the error chasing that we would otherwise be privy to. This is a good application of Theorem 1, as we prove that this method should produce estimates of the response that come arbitrarily close to the true response, given that the response is a continuous function. As a comparison, we see that generally the tree-based methods perform comparably within a particular simulation, but performance changes across different simulations. Depending on the scenario, it seems that multivariate methods are capable of outperforming functional methods with respect to rMSPE. However, one exception to this is the functional neural networks introduced here; they seem to be consistently good performers across these scenarios. With respect to outliers, we can see that they are most prevalent in Simulation 2; this is because this simulation was for when the link function $g(\cdot)$ was exponential. In this context, we expect that the difference between our prediction and the corresponding observed value is greater than it would be in the other simulation scenarios. 

\section{Conclusions and Discussion}
\label{sec:conc}




The extreme rise in popularity of deep learning research has resulted in enormous breakthroughs in computer vision, classification, and scalar prediction. However, these advantages thus far have been limited to when the data is treated as discrete. This paper introduces the first of a family of neural networks that extend into the functional space.

In particular, we present a functional feed-forward neural network to predict a scalar response with functional and scalar covariates. We developed a methodology which showed the steps required to compute the functional weight for the neural network. Multiple examples were provided which showed that the functional neural network outperformed a number of other functional models and multivariate methods with respect to the mean squared prediction error. It was also shown through simulation studies that the recovery of the true functional weight is better done by the functional neural network than the functional linear model when the true relationship is non-linear.


To extend this project, algorithms can be developed for other combinations of input and output types such as the function on function regression models \citep{morris2015functional}. Moreover, one can consider adding additional constraints to the first-layer neurons via penalization or other methods.

\newpage

\begin{center}
{\large\bf SUPPLEMENTAL MATERIALS}
\end{center}

\begin{description}

\item[Data and R Codes:] Data and R Codes used for each application and simulation study presented in this paper; this includes the source code for the upcoming R package. A README file is included which describes each file. (FNNcode.zip File)

\item[Supplemental Document:] Document containing the proof for Theorem 1, a table with descriptions of various parameters, model configurations, and additional simulation results. (supplementalFNN.pdf File)

\end{description}

\bibliography{references}

\begin{thebibliography}{34}
\newcommand{\enquote}[1]{``#1''}
\expandafter\ifx\csname natexlab\endcsname\relax\def\natexlab#1{#1}\fi

\bibitem[{Aneiros-P{\'e}rez and Vieu(2006)}]{aneiros2006semi}
Aneiros-P{\'e}rez, G.,  and Vieu, P. (2006), \enquote{Semi-functional partial
  linear regression,} \textit{Statistics \& Probability Letters}, 76,
  1102--1110.

\bibitem[{Breiman(2001)}]{breiman2001random}
Breiman, L. (2001), \enquote{Random forests,} \textit{Machine learning}, 45,
  5--32.

\bibitem[{Cardot et~al.(1999)Cardot, Ferraty, and Sarda}]{cardot1999functional}
Cardot, H., Ferraty, F.,  and Sarda, P. (1999), \enquote{Functional linear
  model,} \textit{Statistics \& Probability Letters}, 45, 11--22.

\bibitem[{Chen et~al.(2015)Chen, He, Benesty, Khotilovich, and
  Tang}]{chen2015xgboost}
Chen, T., He, T., Benesty, M., Khotilovich, V.,  and Tang, Y. (2015),
  \enquote{Xgboost: extreme gradient boosting,} \textit{R package version
  0.4-2}, 1--4.

\bibitem[{Chollet et~al.(2015)}]{chollet2015keras}
Chollet, F. et~al. (2015), \enquote{Keras,} \url{https://keras.io}.

\bibitem[{Cybenko(1989)}]{cybenko1989approximation}
Cybenko, G. (1989), \enquote{Approximation by superpositions of a sigmoidal
  function,} \textit{Mathematics of Control, Signals and Systems}, 2, 303--314.

\bibitem[{Fanaee-T and Gama(2014)}]{fanaee2014event}
Fanaee-T, H.,  and Gama, J. (2014), \enquote{Event labeling combining ensemble
  detectors and background knowledge,} \textit{Progress in Artificial
  Intelligence}, 2, 113--127.

\bibitem[{Febrero-Bande and de~la Fuente(2012)}]{febrero2012statistical}
Febrero-Bande, M.,  and de~la Fuente, M. (2012), \enquote{Statistical Computing
  in Functional Data Analysis: The R Package fda.usc,} \textit{Journal of
  Statistical Software}, 51, 1--28.

\bibitem[{Fern{\'a}ndez-Redondo and
  Hern{\'a}ndez-Espinosa(2001)}]{fernandez2001weight}
Fern{\'a}ndez-Redondo, M.,  and Hern{\'a}ndez-Espinosa, C. (2001),
  \enquote{Weight initialization methods for multilayer feedforward.} in
  \textit{European Symposium on Artificial Neural Networks}.

\bibitem[{Ferraty and Vieu(2006)}]{ferraty2006nonparametric}
Ferraty, F.,  and Vieu, P. (2006), \textit{Nonparametric functional data
  analysis: theory and practice}, New York: Springer-Verlag.

\bibitem[{Friedman(2001)}]{friedman2001greedy}
Friedman, J. (2001), \enquote{Greedy function approximation: a gradient
  boosting machine,} \textit{Annals of Statistics}, 29, 1189--1232.

\bibitem[{Friedman and Stuetzle(1981)}]{friedman1981projection}
Friedman, J.,  and Stuetzle, W. (1981), \enquote{Projection pursuit
  regression,} \textit{Journal of the American Statistical Association}, 76,
  817--823.

\bibitem[{Hahnloser et~al.(2000)Hahnloser, Sarpeshkar, Mahowald, Douglas, and
  Seung}]{hahnloser2000digital}
Hahnloser, R.~H., Sarpeshkar, R., Mahowald, M.~A., Douglas, R.~J.,  and Seung,
  H.~S. (2000), \enquote{Digital selection and analogue amplification coexist
  in a cortex-inspired silicon circuit,} \textit{Nature}, 405, 947--951.

\bibitem[{Han and Moraga(1995)}]{han1995influence}
Han, J.,  and Moraga, C. (1995), \enquote{The influence of the sigmoid function
  parameters on the speed of backpropagation learning,} in
  \textit{International Workshop on Artificial Neural Networks}, Springer.

\bibitem[{He et~al.(2016)He, Zhang, Ren, and Sun}]{he2016deep}
He, K., Zhang, X., Ren, S.,  and Sun, J. (2016), \enquote{Deep residual
  learning for image recognition,} in \textit{Proceedings of the IEEE
  Conference on Computer Vision and Pattern Recognition}.

\bibitem[{Jiang and Wang(2011)}]{jiang2011functional}
Jiang, C.-R.,  and Wang, J.-L. (2011), \enquote{Functional single index models
  for longitudinal data,} \textit{The Annals of Statistics}, 39, 362--388.

\bibitem[{Kim and Ra(1991)}]{kim1991weight}
Kim, Y.,  and Ra, J. (1991), \enquote{Weight value initialization for improving
  training speed in the backpropagation network,} in \textit{IEEE International
  Joint Conference on Neural Networks}, IEEE.

\bibitem[{Kingma and Ba(2014)}]{kingma2014adam}
Kingma, D.,  and Ba, J. (2014), \enquote{Adam: A method for stochastic
  optimization,} \textit{arXiv preprint arXiv:1412.6980}.

\bibitem[{Krizhevsky et~al.(2012)Krizhevsky, Sutskever, and
  Hinton}]{krizhevsky2012imagenet}
Krizhevsky, A., Sutskever, I.,  and Hinton, G.~E. (2012), \enquote{Imagenet
  classification with deep convolutional neural networks,} in \textit{Advances
  in Neural Information Processing Systems}.

\bibitem[{Morris(2015)}]{morris2015functional}
Morris, J.~S. (2015), \enquote{Functional regression,} \textit{Annual Review of
  Statistics and Its Application}, 2, 321--359.

\bibitem[{M{\"u}ller and Stadtm{\"u}ller(2005)}]{muller2005generalized}
M{\"u}ller, H.-G.,  and Stadtm{\"u}ller, U. (2005), \enquote{Generalized
  functional linear models,} \textit{The Annals of Statistics}, 33, 774--805.

\bibitem[{Preda et~al.(2007)Preda, Saporta, and L{\'e}v{\'e}der}]{preda2007pls}
Preda, C., Saporta, G.,  and L{\'e}v{\'e}der, C. (2007), \enquote{PLS
  classification of functional data,} \textit{Computational Statistics}, 22,
  223--235.

\bibitem[{Ramsay et~al.(2009)Ramsay, Hooker, and Graves}]{fda_matlab}
Ramsay, J.~O., Hooker, G.,  and Graves, S. (2009), \textit{Functional data
  analysis with R and MATLAB}, New York: Springer.

\bibitem[{Ramsay and Silverman(2005)}]{Ramsay05}
Ramsay, J.~O.,  and Silverman, B.~W. (2005), \textit{Functional Data Analysis},
  New York: Springer.

\bibitem[{Rossi and Conan-Guez(2005)}]{rossi2005functional}
Rossi, F.,  and Conan-Guez, B. (2005), \enquote{Functional multi-layer
  perceptron: a non-linear tool for functional data analysis,} \textit{Neural
  Networks}, 18, 45--60.

\bibitem[{Ruder(2016)}]{ruder2016overview}
Ruder, S. (2016), \enquote{An overview of gradient descent optimization
  algorithms,} \textit{arXiv preprint arXiv:1609.04747}.

\bibitem[{Rumelhart et~al.(1985)Rumelhart, Hinton, and
  Williams}]{rumelhart1985learning}
Rumelhart, D., Hinton, G.,  and Williams, R. (1985), \enquote{Learning internal
  representations by error propagation,} Tech. rep., California Univ San Diego
  La Jolla Inst for Cognitive Science.

\bibitem[{Seber and Lee(2012)}]{seber2012linear}
Seber, G.~A.,  and Lee, A.~J. (2012), \textit{Linear regression analysis},
  Hoboken: John Wiley \& Sons.

\bibitem[{Srivastava et~al.(2014)Srivastava, Hinton, Krizhevsky, Sutskever, and
  Salakhutdinov}]{srivastava2014dropout}
Srivastava, N., Hinton, G., Krizhevsky, A., Sutskever, I.,  and Salakhutdinov,
  R. (2014), \enquote{Dropout: a simple way to prevent neural networks from
  overfitting,} \textit{Journal of Machine Learning Research}, 15, 1929--1958.

\bibitem[{Süli and Mayers(2003)}]{Suli03}
Süli, E.,  and Mayers, D. (2003), \textit{An Introduction to Numerical
  Analysis}, Cambridge: Cambridge University Press.

\bibitem[{Thodberg(2015)}]{thodberg2015tecator}
Thodberg, H.~H. (2015), \enquote{Tecator meat sample dataset,} \textit{StatLib
  Datasets Archive}.

\bibitem[{Tibshirani(1996)}]{tibshirani1996regression}
Tibshirani, R. (1996), \enquote{Regression shrinkage and selection via the
  lasso,} \textit{Journal of the Royal Statistical Society: Series B
  (Methodological)}, 58, 267--288.

\bibitem[{Tibshirani et~al.(2009)Tibshirani, Hastie, and Friedman}]{ESL}
Tibshirani, R., Hastie, T.,  and Friedman, J. (2009), \textit{The elements of
  statistical learning: data Mining, inference, and prediction}, New York:
  Springer.

\bibitem[{Yao et~al.(2007)Yao, Rosasco, and Caponnetto}]{yao2007early}
Yao, Y., Rosasco, L.,  and Caponnetto, A. (2007), \enquote{On early stopping in
  gradient descent learning,} \textit{Constructive Approximation}, 26,
  289--315.

\end{thebibliography}

\end{document}